\documentclass[11pt]{article}
\usepackage[preprint]{acl}
\usepackage{times}
\usepackage{latexsym}
\usepackage[T1]{fontenc}
\usepackage[utf8]{inputenc}
\usepackage{microtype}
\usepackage{amsmath,amssymb}
\usepackage{booktabs}
\usepackage{graphicx}
\usepackage{xcolor}

\newcommand{\hEntries}{860}
\newcommand{\hEntriesYI}{535}
\newcommand{\hEntriesYII}{325}
\newcommand{\hWords}{203{,}819}
\newcommand{\hPages}{394}

\newcommand{\hAccRate}{82.4}
\newcommand{\hSilver}{13{,}122}

\newcommand{\hTrain}{10{,}525}
\newcommand{\hDev}{1{,}313}

\newcommand{\hTestLines}{2{,}004}
\newcommand{\hAlignPrec}{99.98}

\newcommand{\hSigla}{303}

\newcommand{\hNoiseLow}{0.13}
\newcommand{\hNoisePlaus}{0.46}

\newcommand{\hCepochs}{57}
\newcommand{\hCbest}{51}
\newcommand{\hCsec}{54}
\newcommand{\hCmin}{52}
\newcommand{\hCdev}{5.89}
\newcommand{\hBstages}{6}
\newcommand{\hBbest}{3}
\newcommand{\hBacc}{92.3}
\newcommand{\hCcer}{7.1}
\newcommand{\hCline}{7.8}
\newcommand{\hCpage}{15.9}
\newcommand{\hCwer}{14.7}
\newcommand{\hBcer}{8.3}
\newcommand{\hBwer}{18.3}

\newcommand{\hAcmp}{18.9}
\newcommand{\hAexp}{18.0}
\newcommand{\hCfifteen}{7.3}
\newcommand{\hPageTwentyEight}{17.4}
\newcommand{\hCmpTwentyEight}{6.8}
\newcommand{\hSubTenWer}{28.9}
\newcommand{\hSubTenCer}{11.5}

\newcommand{\hShareAbbr}{47}
\newcommand{\hErrAbbr}{14.4}
\newcommand{\hErrFull}{15.0}
\newcommand{\hErrFinal}{22.3}

\newcommand{\hOovWer}{26.9}
\newcommand{\hInvWer}{13.6}

\newcommand{\hBoundaryShare}{2.0}
\newcommand{\hSegF}{0.994}

\newcommand{\hEdWords}{254{,}716}

\newcommand{\hSFepochs}{60}
\newcommand{\hSiglaOld}{10.2}
\newcommand{\hSiglaNew}{5.4}

\newcommand{\hAurocIso}{0.870}

\newcommand{\hEce}{0.8}

\newcommand{\hCovA}{74}
\newcommand{\hCovC}{82}
\newcommand{\hCovCN}{1{,}648}
\newcommand{\hCovBest}{86}
\newcommand{\hCovCharC}{94}

\newcommand{\hClauseEnds}{600}
\newcommand{\hCanonShare}{71}
\newcommand{\hLmExcl}{263}
\newcommand{\hLmEntries}{860}
\newcommand{\hLmExclNoFolio}{21}
\newcommand{\hPadMax}{810}

\newcommand{\hPadMaskInf}{8}

\newcommand{\hHybWer}{12.5}
\newcommand{\hHybCer}{6.4}
\newcommand{\hHybFinal}{16.4}
\newcommand{\hHybOov}{24.3}
\newcommand{\hHybAlpha}{8.1}
\newcommand{\hHybBeta}{10.1}
\newcommand{\hCFinal}{22.7}
\newcommand{\hMaskWer}{14.3}

\newcommand{\hPtrOov}{24.9}
\newcommand{\hParWer}{15.0}
\newcommand{\hDiplWer}{15.1}
\newcommand{\hMaeWer}{14.7}
\newcommand{\hLMwerL}{13.0}
\newcommand{\hSFwerL}{15.6}
\newcommand{\hSFwerOneL}{20.8}
\newcommand{\hReSegWerL}{16.2}
\newcommand{\hCwerL}{14.7}
\newcommand{\hMaeTenWer}{29.4}
\newcommand{\hScrTenWer}{28.0}

\newcommand{\hnEWerExp}{12.5}
\newcommand{\hnECerExp}{6.4}
\newcommand{\hnEWerSig}{8.9}

\newcommand{\hnEWerCh}{13.3}
\newcommand{\hnECerCh}{10.1}
\newcommand{\hnENames}{32.1}

\newcommand{\hnVWerExp}{12.8}
\newcommand{\hnVCerExp}{6.8}
\newcommand{\hnVWerSig}{8.7}

\newcommand{\hnVWerCh}{9.2}
\newcommand{\hnVCerCh}{5.0}
\newcommand{\hnVNames}{29.6}

\newcommand{\hnNWerExp}{12.8}
\newcommand{\hnNCerExp}{6.8}
\newcommand{\hnNWerSig}{8.7}
\newcommand{\hnNCerSig}{4.3}
\newcommand{\hnNWerCh}{8.2}
\newcommand{\hnNCerCh}{4.9}

\newcommand{\hnNAlpha}{7.2}
\newcommand{\hnNBeta}{10.6}

\newcommand{\hnChangedLines}{31}
\newcommand{\hnTagged}{630}

\newcommand{\hnHeadFone}{0.74}
\newcommand{\hnHeadP}{0.78}
\newcommand{\hnHeadR}{0.70}
\newcommand{\hnEFinal}{16.4}

\newcommand{\hnNFinal}{17.3}

\newcommand{\hnEtoEF}{0.69}
\newcommand{\hnEtoEP}{0.72}
\newcommand{\hnEtoER}{0.66}
\newcommand{\hnFilterF}{0.78}
\newcommand{\hnOrF}{0.73}
\newcommand{\hnAndP}{0.85}

\newcommand{\hWkCvTwoWer}{13.2}
\newcommand{\hWkCvTwoCer}{5.8}

\newcommand{\hWkClmWer}{9.7}

\newcommand{\hWkHvTwoWer}{11.0}
\newcommand{\hWkHvTwoCer}{5.3}
\newcommand{\hWkHvTwoFinal}{15.7}

\newcommand{\hWkRtlWer}{10.6}
\newcommand{\hWkRtlCer}{5.1}
\newcommand{\hWkRtlFinal}{13.7}

\newcommand{\hWkSeedWer}{10.6}
\newcommand{\hWkSeedCer}{5.4}
\newcommand{\hWkSeedFinal}{14.6}

\newcommand{\hWkBestWer}{8.7}
\newcommand{\hWkBestCer}{4.7}
\newcommand{\hWkBestFinal}{12.2}

\newcommand{\hWkBestAlpha}{7.9}
\newcommand{\hWkBestBeta}{5.8}
\newcommand{\hWkRovWer}{8.4}

\newcommand{\hWkRovFinal}{11.5}
\newcommand{\hWkRovOov}{22.0}

\newcommand{\hWkNuWerCh}{8.2}
\newcommand{\hWkNuCerCh}{5.3}

\newcommand{\hWkNbWer}{9.8}

\newcommand{\hWkNbWerCh}{7.7}
\newcommand{\hWkNbCerCh}{5.2}

\newcommand{\hWkNmWerCh}{7.7}
\newcommand{\hWkNmCerCh}{5.7}

\newcommand{\hWkNsWer}{9.6}

\newcommand{\hWkNbestWer}{9.8}
\newcommand{\hWkNbestCer}{5.1}
\newcommand{\hWkNbestWerSig}{7.2}
\newcommand{\hWkNbestCerSig}{3.8}
\newcommand{\hWkNbestWerCh}{7.7}

\newcommand{\hWkMaskCer}{6.9}

\newcommand{\hWkSilverTwo}{12{,}725}
\newcommand{\hWkBestDevWer}{7.45}

\newcommand{\hWkMbrDevWer}{7.49}

\newcommand{\hPadHybOld}{402}
\newcommand{\hPadHybFixed}{3}
\newcommand{\hPadBpeFixed}{1}
\newcommand{\hPadHybOldWerLo}{11.62}
\newcommand{\hPadHybOldWerHi}{12.25}

\newcommand{\hBsCWerD}{$-$4.6}
\newcommand{\hBsCWerLo}{$-$5.2}
\newcommand{\hBsCWerHi}{$-$4.0}

\newcommand{\hBsHybWerD}{$-$3.7}
\newcommand{\hBsHybWerLo}{$-$4.3}
\newcommand{\hBsHybWerHi}{$-$3.2}

\newcommand{\hBsHybCerD}{$-$1.7}
\newcommand{\hBsHybCerLo}{$-$2.1}
\newcommand{\hBsHybCerHi}{$-$1.3}

\newcommand{\hBsRovWerD}{$-$4.0}
\newcommand{\hBsRovWerLo}{$-$4.7}
\newcommand{\hBsRovWerHi}{$-$3.5}

\newcommand{\hBsNerWerD}{$-$3.1}

\newcommand{\hBsNerSigWerD}{$-$1.5}

\newcommand{\hBsNerChWerLo}{$-$2.7}
\newcommand{\hBsNerChWerHi}{$+$1.2}

\newcommand{\hBsSeedWerD}{$-$1.9}
\newcommand{\hBsSeedWerLo}{$-$2.4}
\newcommand{\hBsSeedWerHi}{$-$1.3}

\newcommand{\hBsTgtWerD}{$+$0.3}

\newcommand{\hBsTgtCerD}{$+$0.4}

\newcommand{\hBsTgtChWerD}{$-$4.1}
\newcommand{\hBsTgtChWerLo}{$-$6.3}
\newcommand{\hBsTgtChWerHi}{$-$1.7}

\newcommand{\hBsTgtChCerD}{$-$5.1}
\newcommand{\hBsTgtChCerLo}{$-$7.0}
\newcommand{\hBsTgtChCerHi}{$-$3.3}

\newcommand{\hBsHeadChWerD}{$-$1.0}
\newcommand{\hBsHeadChWerLo}{$-$2.5}
\newcommand{\hBsHeadChWerHi}{$+$0.7}

\title{Reading the Registers of Honorius~III without Manual Line Transcription: Forced-Aligned HTR, End to End, Modular or Hybrid}

\author{Tam\'as Kov\'acs \\
  University of Graz \\
  Department of Digital Humanities \\\And
  Daniel Luger \\
  Institut f\"ur \"Osterreichische Geschichtsforschung \\
  University of Vienna}

\begin{document}
\maketitle

\begin{abstract}
Unlocking medieval archives at scale is bottlenecked by the cost of line-level diplomatic transcriptions and manual abbreviation expansion. Focusing on the registers of Pope Honorius~III (1216--1227)---over 5,000 letters surviving in full only as manuscript images---we show that existing, line-break-free expert transcriptions (edition drafts) suffice to train end-to-end handwritten text recognition (HTR) that jointly transcribes and expands heavily abbreviated Latin. Using forced alignment on \hEntries{} unaligned edition entries across \hPages{} pages of Reg.\ Vat.~9, we automatically bootstrap \hSilver{} training lines without manual alignment of segmented lines. While the generic CATMuS baseline struggles on this domain (\hAcmp\% CER) and a modular space-free HTR $+$ word-segmentation pipeline (cf.\ \citealt{camps2021}) plateaus at \hSFwerL\% WER, our from-scratch hybrid Conformer---combining CTC with bidirectional subword attention decoders, iterative self-realignment and $n$-best LM rescoring---achieves \hWkBestCer\% CER and \hWkBestWer\% WER (\hWkRovWer\% WER with ROVER system combination), nearly halving errors on clause-final words (\hCFinal\% $\rightarrow$ \hWkBestFinal\% WER). These figures are measured on the test lines that align with the transcription (\hCovBest\% of all test lines for the best hybrid). Furthermore, we address a tension between paleographic fidelity and editorial normalization: training on targets that preserve scribal name initials reduces the WER against the manuscript text on the \hnChangedLines{} affected test lines from \hnEWerCh\% to \hnVWerCh\%, at a small cost against the expanded transcription (WER \hBsTgtWerD{} points, CER \hBsTgtCerD{}). Finally, extensive ablations show that none of an auxiliary diplomatic CTC head, a copy mechanism, a permutation decoder or masked-image pretraining improves on direct end-to-end supervision, while strict padding-mask propagation and per-line beam termination make inference batch-invariant.
\end{abstract}

\section{Introduction}

For whole regions of thirteenth-century history the papal registers are the most important source, yet they remain incompletely edited and are often misread \citep{schabel2023}. The registers of Honorius~III (Reg.\ Vat.~9--13) have no published edition, and digital methods from text reuse to the study of rhythmic prose, such as the cursus of clause endings, presuppose a machine-readable text \citep{vogeler2023}. Two features of the source make HTR hard. In the dense curial minuscule, segmentation and recognition depend on each other \citep{sayre1973}. Abbreviation is pervasive: \hShareAbbr\% of the words of Reg.\ Vat.~9 are abbreviated, as in contemporary scholastic hands \citep{camps2021}. A graphemic transcription that keeps the signs, as in CATMuS \citep{clerice2024}, leaves case endings and syllable counts undetermined. Historical analysis needs the expanded word, but expanded line-level ground truth is scarce.

Earlier HTR of these registers worked around this scarcity. In Codice Ratio (ICR) trained a character classifier on crowd-sourced labels and decoded with a language model \citep{firmani2018,ammirati2019}. On 39 pages of Reg.\ Vat.~12 the crowd-enabled pipeline reached 19.18\% CER, and a Transkribus model built on 9,000 expert lines of other material 15.90\% \citep{nieddu2021}. ICR then trained on whole lines transcribed by experts on about 20 folios. The resulting model transcribed Reg.\ Vat.~12 with at most eight errors per 80-character line; its residual errors were unseen abbreviations and inconsistent expansions in the targets \citep{ammirati2023,ammirati2025}. For other chancery registers, ground truth has been obtained by aligning images with existing editions \citep{stutzmann2017,torresaguilar2023}. Generic models trained on expanded documentary text replicate the editorial convention of their training data \citep{torresaguilar2023}. Abbreviation and word boundary are independent choices of the scribe \citep{hasenohr1998}, and word segmentation can be solved as character tagging \citep{clerice2020}. On a heavily abbreviated codex with 1,524 training lines, \citet{camps2021} found that a modular chain was better than unsegmented expanded output. In their chain, diplomatic recognition is followed by a segmenter and a normalizer, and the chain reached 31.81\% WER against 54.68\%. Whether this holds with more expanded data has not been tested on one corpus.

We contribute (i) a forced-alignment route from existing transcriptions to \hSilver{} silver lines, (ii) expansion-aware recognizers with an accuracy that improves on earlier work on the registers, and (iii) a same-corpus comparison of end-to-end, modular and hybrid recognition, with segmenters and language models trained on papal Latin, attention decoders, self-supervised pretraining and a batch-invariant encoder, (iv) targets that keep name initials as written, with and without a jointly trained named-entity head, which show that the target convention, not the head, decides whether sigla are reproduced, and (v) iterative re-alignment with bidirectional and subword decoding, which lowers the WER of the hybrid by 30\%.

\begin{table*}[t]
\centering\small\setlength{\tabcolsep}{3.5pt}
\resizebox{\textwidth}{!}{%
\begin{tabular}{llllll}
\toprule
Study & Source & Training data & Test & Target & Result \\
\midrule
ICR crowd pipeline \citep{nieddu2021} & Reg.\ Vat.~12 & 20 crowd-annotated pages & 39 pages & letters + abbr.\ symbols & CER 19.18\% \\
Transkribus model \citep{nieddu2021} & Reg.\ Vat.~12 & 9,000 expert lines, other material & same 39 pages & -- & CER 15.90\% \\
ICR expert lines \citep{ammirati2023,ammirati2025} & Reg.\ Vat.~12 & $\approx$20 folios & 5 folios & expanded & $\le$8 errors / 80-char.\ line \\
CATMuS, Kraken \citep{clerice2024} & 8th--16th c. & 146,885 / 138,900 lines & seen / held-out mss. & graphemic & CER 4.7 / 13.1\% \\
\citet{camps2021}, full systems & BnF lat.~14525 & 1,524 lines, one scribe & folio 45v & expanded & CER 6.89--8.60\%, WER 31.81--54.68\% \\
\citet{torresaguilar2023} & charters, registers & $\approx$105,000 lines & held-out mss. & semi-diplomatic & accuracy $\ge$0.85 / $\ge$0.90 \\
\midrule
A: CATMuS Medieval 1.5 & Reg.\ Vat.~9 & -- & 39 pages & graphemic & CER \hAcmp\% \\
B: CATMuS fine-tuned & Reg.\ Vat.~9 & \hTrain{} silver lines & same 39 pages & expanded & CER \hBcer\%, WER \hBwer\% \\
C: Conformer-CTC & Reg.\ Vat.~9 & \hTrain{} silver lines & same 39 pages & expanded & CER \hCcer\%, WER \hCwer\% \\
Hybrid CTC/attention & Reg.\ Vat.~9 & \hTrain{} silver lines & same 39 pages & expanded & CER \hHybCer\%, WER \hHybWer\% \\
Hybrid + NER head & Reg.\ Vat.~9 & \hTrain{} silver lines, sigla as written & same 39 pages & expanded / sigla as written & CER \hnNCerExp\% / \hnNCerSig\%, WER \hnNWerExp\% / \hnNWerSig\% \\
Hybrid, re-aligned, R2L + BPE + LM & Reg.\ Vat.~9 & \hWkSilverTwo{} re-aligned lines & same 39 pages & expanded & CER \hWkBestCer\%, WER \hWkBestWer\% \\
Hybrid + NER head, re-aligned, R2L + LM & Reg.\ Vat.~9 & \hWkSilverTwo{} re-aligned lines, sigla as written & same 39 pages & expanded / sigla as written & CER \hWkNbestCer\% / \hWkNbestCerSig\%, WER \hWkNbestWer\% / \hWkNbestWerSig\% \\
\bottomrule
\end{tabular}}
\caption{Published results on the papal registers and related material, and the recognizers of this study (held-out pages of the same register). Test sets, targets and metrics differ; the comparison is indicative.}
\label{tab:prior}
\end{table*}

\section{Data and Silver Lines}

Reg.\ Vat.~9 contains the first regnal years of Honorius~III. We use \hPages{} page images (fol.~1--130 and 137--203) and two expert transcriptions: the edition draft of regnal year~1 by D.~Luger (\hEntriesYI{} entries) and a working transcription of the first half of year~2 by A.~Singh Bottanova (\hEntriesYII{} entries), \hWords{} Latin words in total. Both expand abbreviations and have no line breaks. Apparatus marks and editorial notes are removed, and parenthesized expansions are joined to their word.

Kraken \citep{kiessling2019} segments each page, and CATMuS Medieval~1.5 \citep{clerice2024} reads every line. Each line is located in the transcription of the entries on its folio by semi-global edit distance on a harmonized reading layer (lower case; no punctuation; u/v, i/j, ae/e, ti/ci and MUFI signs normalized). The search re-anchors when a page begins in mid-entry. The matched span, extended to word boundaries and in the original spelling, becomes the line's target. Lines with normalized distance at most 0.35 are accepted (\hAccRate\%); those of at least 15 characters with distance at most 0.25 form the silver set (\hSilver{} lines; Appendix~\ref{app:data}). In simulation, \hAlignPrec\% of accepted lines are placed correctly at 15\% CER with half of the words abbreviated. Test and development pages are regularly interleaved with the training pages: the test pages are the versos of folios 3, 8, 13, \dots{} and the development pages the rectos of folios 1, 6, 11, \dots{} (every tenth page each), so the test set consists of versos only and the development set of rectos only. This gives 315 / 40 / 39 pages and \hTrain{} training and \hDev{} development lines.

The targets inherit the conventions of the transcriptions. Editorial expansions of name initials reach them (\hSigla{} tokens, e.g.\ \emph{Berengarie} for a bare \emph{B.}); footnote marks and editorial notes do not. A lexicon check puts transcription slips at \hNoiseLow--\hNoisePlaus\% of the reference words.

\section{Recognizers and Variants}

\textbf{A} is CATMuS Medieval~1.5 as published; it keeps abbreviation signs. \textbf{B} fine-tunes A on the silver lines with Kraken's trainer. \textbf{C} is a Conformer-CTC \citep{gulati2020,graves2006} trained from scratch: a four-block convolutional front-end on 64-pixel line images, eight Conformer blocks ($d=256$) and greedy CTC decoding (12.5\,M parameters, about \hCmin{} minutes on one GPU). All variants share data, split, seed, optimizer and early stopping (Appendix~\ref{app:training}).

All Conformer models propagate the padding mask through every convolution, pooling and batch-normalization layer, and the attention beam search stops each line at its own $K$ finished hypotheses (Appendix~\ref{app:repro}); \emph{masked C} is C retrained with this masking. \emph{LM fusion} adds a Kneser--Ney word trigram model, trained on the Innocent~III edition (\hEdWords{} words) and on the Honorius entries that do not touch a development or test page, to CTC prefix beam search. An entry is left out if its folio range includes any development or test page, so an entry that only partly lies on a held-out page is excluded as well, as is every entry without a folio assignment (\hLmExcl{} of \hLmEntries{} entries, \hLmExclNoFolio{} of them without folio). The character and BPE models below are trained on the same text. Four \emph{attention decoders} (four-layer Transformers, trained jointly with the CTC head: 0.3\,CTC $+$ 0.7 cross-entropy) sit on the masked encoder: the plain \emph{hybrid} (5-best beam rescored by CTC), a \emph{copy gate} from the CTC distribution, a PARSeq-style \emph{permutation decoder}, and a \emph{diplomatic head} with CTC on the noisy CATMuS reading. \emph{Masked-image pretraining} reconstructs hidden column spans of the training lines. The \emph{modular route} of \citet{camps2021} retrains C without spaces and restores word boundaries with a character tagger \citep{clerice2020} trained on Honorius text (S1), with the Innocent~III edition (S2), or with a word bigram model in addition (S3). \emph{Sigla as written} retrains C on targets that keep name initials as written.

\paragraph{NER head.} The transcription expands the initial \emph{.B.} to \emph{Berengarie}, letters not on the page. We fine-tune the hybrid on sigla-as-written targets alone (control) and with a \emph{named-entity head}, a linear layer labeling every output character (O, B/I-person, B/I-place) on rule-derived labels, trained jointly (0.3\,CTC $+$ 0.7 cross-entropy $+$ 0.5 NER; Appendix~\ref{app:names}). At decoding, a word tagged as a name may take a siglum from the 5-best list.

\paragraph{Iterative alignment and subword/bidirectional decoding.} We re-align the 315 training pages with the expanded readings of the hybrid (C where it does not align) instead of the CATMuS reading, and cut words split across lines between stem and ending. The readings only locate spans of the expert transcription, giving \hWkSilverTwo{} training lines (\emph{silver v2}); development and test lines are unchanged. On silver v2 we retrain C with rotary position embeddings and depthwise kernels of 7 and 23 (\emph{RoPE/7-23}), and the hybrid with a second, right-to-left decoder (\emph{R2L}; 0.3\,CTC $+$ 0.5 left-to-right $+$ 0.2 right-to-left) and with decoder targets in 512 byte-pair units (\emph{BPE-512}). Character 10-gram and BPE 4-gram models, trained on the text of the word model, enter C's beam search (\emph{char/BPE LM}) or rescore a 16-best list of the hybrid (\emph{LM rescoring}). The \emph{guarded ROVER} takes the best single hybrid (R2L, BPE-512, LM rescoring) as primary and replaces one of its words only where both of two other systems propose the same different word, one-to-one: C on silver v2 with RoPE/7-23 and char/BPE LM fusion, and the R2L hybrid on silver v2 with LM rescoring but character targets. For the NER hybrid, a \emph{sigla guard} withholds from the language model one-letter initials and the words aligned with them or tagged as names by the head, so that the recognizer's own score decides between a siglum and a longer form. All weights and choices are fixed on the development pages.

Error rates are Levenshtein distances on the reading layer against the expanded transcription. CER (accepted lines $\ge$15 characters) is the mean of page CERs in all tables; WER counts reference words not reproduced, aligned per line. Significance: paired bootstrap over the 39 test pages (10,000 resamples) on pooled rates, which differ from page means by a few tenths at most (other bases: Appendix~\ref{app:bases}).

\paragraph{Coverage of the headline figures.} CER and WER are measured on the lines that the alignment accepts (distance $\le0.35$) and whose reference span has at least 15 characters. For C these are \hCovCN{} of the \hTestLines{} test lines (\hCovC\%; \hCovCharC\% of the recognized characters), for the best hybrid \hCovBest\% and for CATMuS \hCovA\%. The set depends on the recognizer, because the alignment uses its reading; the lines left out are short, marginal or too badly read to align. The page CER, which compares all recognized lines of a page's main text block with the reference and so also counts marginalia, rubrics and reading-order errors, is \hCpage\% for C against \hCcer\% on the aligned lines (Appendix~\ref{app:bases}). The headline figures therefore describe the lines that can be aligned with the transcription, not the whole page.

\paragraph{Cursus.} The cursus is the rhythm of the last words of a clause in medieval Latin prose, defined by the positions of the last two word stresses; its canonical forms are \emph{planus} (\emph{illum deduxit}), \emph{tardus} (\emph{ire tentaverit}) and \emph{velox} (\emph{hominem recepistis}), and its distribution differs between chanceries and periods \citep{janson1975}. Whether a recognizer keeps it matters for every study of papal style that uses the machine-read text. We assign the form of a clause end (. ? ; !, excluding abbreviation points) to the last words with a rule-based classifier that stresses the words with a lexicon and rules and reads the pattern, on the expert transcription and on the recognizer's output. The \hClauseEnds{} clause ends of the test pages are canonical in \hCanonShare\% of the cases in the expert text. $\alpha$ is the share of the clause ends that are not canonical in the expert text but are read as canonical (a cursus that is not there), and $\beta$ the share of canonical clause ends that are read as non-canonical (a cursus that is lost). A fall of $\beta$ from \hHybBeta\% to \hWkBestBeta\% thus means that fewer cursus endings are destroyed by misread final words.

\section{Results}

\paragraph{Accuracy and earlier work.} Table~\ref{tab:prior} compares with published results (indicatively: registers, test sets and metrics differ). Learning expansion from expanded targets is decisive: C reaches \hCcer\% CER and \hCwer\% WER, B \hBcer\% and \hBwer\%, while rule-based expansion of A's signs lowers its line CER only to \hAexp\%. Data volume explains most of the gain: with 1,052 training lines, about the set of \citet{camps2021}, C reaches \hSubTenCer\% CER and \hSubTenWer\% WER, and the learning curve has not flattened at 10,525 lines (Appendix~\ref{app:curve}). Generic models lose accuracy on unseen manuscripts \citep{clerice2024}.

\begin{table}[!b]
\centering\small\setlength{\tabcolsep}{3pt}
\resizebox{\columnwidth}{!}{\begin{tabular}{lrrrrrr}
\toprule
Recognizer & WER & CER & final & OOV & $\alpha$ & $\beta$ \\
 & \% & \% & \% & \% & \% & \% \\
\midrule
\multicolumn{7}{l}{\emph{Baseline and decoding}} \\
\quad C, greedy CTC & 14.7 & 7.1 & 22.7 & 26.9 & 10.5 & 13.6 \\
\quad C retrained with mask propagation & 14.3 & 6.9 & 19.8 & 26.6 & 8.4 & 11.0 \\
\quad C + word trigram LM & 13.0 & 6.8 & 19.9 & 26.0 & 9.9 & 11.6 \\
\multicolumn{7}{l}{\emph{Attention decoders (on masked C)}} \\
\quad hybrid CTC/attention & 12.5 & 6.4 & 16.4 & 24.3 & 8.1 & 10.1 \\
\quad hybrid + copy gate & 13.1 & 6.8 & 19.3 & 24.9 & 12.0 & 9.1 \\
\quad permutation (PARSeq-style) decoder & 15.0 & 7.9 & 22.7 & 27.0 & 15.9 & 13.4 \\
\quad CTC head on diplomatic reading & 15.1 & 8.4 & 20.5 & 27.5 & 8.6 & 11.9 \\
\multicolumn{7}{l}{\emph{Data and targets}} \\
\quad masked-image pretraining & 14.7 & 6.9 & 21.8 & 27.7 & 11.5 & 13.7 \\
\quad C, sigla as written & 15.2 & 7.4 & 21.8 & 27.4 & 13.8 & 11.9 \\
\quad hybrid, sigla as written & 12.8 & 6.8 & 17.8 & 24.1 & 7.8 & 10.5 \\
\quad hybrid, sigla as written + NER head & 12.8 & 6.8 & 17.3 & 24.5 & 7.2 & 10.6 \\
\multicolumn{7}{l}{\emph{Modular route}} \\
\quad space-free + segmenter S1 & 20.8 & 7.9 & 24.9 & 38.0 & 12.4 & 15.5 \\
\quad space-free + segmenter S2 & 17.0 & 7.6 & 23.1 & 32.7 & 9.2 & 12.1 \\
\quad space-free + segmenter S3 & 15.6 & 7.4 & 22.4 & 32.6 & 9.1 & 11.8 \\
\quad C, spaces removed, S3 & 16.2 & 7.5 & 22.7 & 34.4 & 11.1 & 15.3 \\
\multicolumn{7}{l}{\emph{Iterative alignment and subword/bidirectional decoding}} \\
\quad C, silver v2 & 13.2 & 5.8 & 17.7 & 25.9 & 12.5 & 10.3 \\
\quad C, silver v2 + RoPE/7-23 + char/BPE LM & 9.7 & 5.0 & 15.4 & 23.0 & 10.4 & 7.5 \\
\quad hybrid, silver v2 & 11.0 & 5.3 & 15.7 & 23.9 & 11.4 & 6.8 \\
\quad hybrid, silver v2 + R2L$^{a}$ & 10.6 & 5.1 & 13.7 & 24.3 & 12.4 & 5.9 \\
\quad hybrid, silver v2 + R2L + BPE-512 & 9.4 & 4.8 & 12.5 & 24.2 & 8.3 & 6.8 \\
\quad \quad + LM rescoring & 8.7 & 4.7 & 12.2 & 24.0 & 7.9 & 5.8 \\
\quad guarded ROVER (best hybrid + 2 systems) & 8.4 & 4.6 & 11.5 & 22.0 & 8.5 & 5.4 \\
\bottomrule
\end{tabular}}
\caption{All variants on the 39 test pages (words aligned per line). Final, OOV: WER on clause-final and out-of-vocabulary words; $\alpha$, $\beta$: cadence errors at clause ends (Section~3): non-canonical endings read as canonical ($\alpha$), canonical endings read as non-canonical ($\beta$). Guarded ROVER: best single hybrid with two other systems (Section~3). Hybrids use the batch-invariant beam; last group trained on silver v2. $^{a}$Second seed: WER \hWkSeedWer\%, CER \hWkSeedCer\%, final \hWkSeedFinal\%.}
\label{tab:variants}
\end{table}

\paragraph{Where the errors are.} Abbreviated and fully written words are misread about equally often (\hErrAbbr\% vs \hErrFull\%); the position in the clause matters more. The clause-final word is wrong in \hCFinal\% of cases, abbreviated clause-final words most often (Appendix~\ref{app:errors}). Words outside the training vocabulary, mostly names, are wrong in \hOovWer\% of cases, others in \hInvWer\%, and only \hBoundaryShare\% of the word errors are pure boundary errors. Trained on expanded sigla, C writes \emph{Berengarie} for \emph{B.}: the target convention decides the output. A calibrated word error probability (AUROC \hAurocIso) can route uncertain words to an expert.

\paragraph{Variants.} An internal language model helps most: the hybrid lowers the WER from \hCwerL\% to \hHybWer\%, the clause-final WER from \hCFinal\% to \hHybFinal\% and the OOV WER from \hOovWer\% to \hHybOov\% (Table~\ref{tab:variants}), and cadence misclassification does not rise ($\alpha$ \hHybAlpha\%, $\beta$ \hHybBeta\%). An external word model helps less (\hLMwerL\%), masked retraining slightly (\hMaskWer\%). The copy gate does not help even for unseen names (\hPtrOov\% OOV WER), the diplomatic head inherits the errors of its noisy layer (\hDiplWer\%), our permutation decoder stays behind the hybrid (\hParWer\%), and masked-image pretraining on the training lines themselves adds nothing (\hMaeWer\%; with 10\% of the labels \hMaeTenWer\% against \hScrTenWer\% from scratch).

The modular route we tested (recognition without spaces, then a segmenter) loses, although its segmenters restore word boundaries in clean text with $F_1=\hSegF$, as reported for medieval Latin \citep{clerice2020}. Papal Latin helps the segmenter (\hSFwerOneL\% $\to$ \hSFwerL\% WER), but the chain does not reach C, and re-segmenting C's output makes it worse (\hReSegWerL\%). The segmenter splits unseen names (OOV WER above 32\%). \citet{camps2021} found the reverse for their chain (diplomatic recognition, segmenter, normalizer) with 1,524 lines; we do not test that chain (see Limitations), but with 10,525 lines the end-to-end recognizer has learned word spacing. For C, sigla as written halves the CER on the affected lines (\hSiglaOld\% $\to$ \hSiglaNew\%); for the hybrid the same change lowers it from \hnECerCh\% to \hnVCerCh\% (Table~\ref{tab:nersigla}), in both cases without a clear overall effect.

\paragraph{Iterative alignment, bidirectional and subword decoding.} Re-alignment alone lowers the WER of C from \hMaskWer\% to \hWkCvTwoWer\% (CER \hWkMaskCer\% $\to$ \hWkCvTwoCer\%) and of the hybrid from \hHybWer\% to \hWkHvTwoWer\% (CER \hHybCer\% $\to$ \hWkHvTwoCer\%). With RoPE/7-23 and char/BPE LM fusion C reaches \hWkClmWer\% WER. The right-to-left decoder helps most at clause ends (\hWkHvTwoFinal\% $\to$ \hWkRtlFinal\%); both seeds of this model give \hWkRtlWer\% WER (CER \hWkRtlCer\% and \hWkSeedCer\%). BPE targets and LM rescoring bring a single hybrid to \hWkBestWer\% WER, \hWkBestCer\% CER and \hWkBestFinal\% clause-final WER, the guarded ROVER to \hWkRovWer\% WER and \hWkRovFinal\% clause-final WER; OOV words gain least (\hWkRovOov\%). $\beta$ falls from \hHybBeta\% to \hWkBestBeta\% while $\alpha$ stays at \hWkBestAlpha\%. Paired bootstrap differences against the first hybrid (95\% CI): best single hybrid WER \hBsHybWerD{} [\hBsHybWerLo, \hBsHybWerHi], CER \hBsHybCerD{} [\hBsHybCerLo, \hBsHybCerHi]; ROVER WER \hBsRovWerD{} [\hBsRovWerLo, \hBsRovWerHi]; second R2L seed WER \hBsSeedWerD{} [\hBsSeedWerLo, \hBsSeedWerHi]; new C against masked C WER \hBsCWerD{} [\hBsCWerLo, \hBsCWerHi]; all $p<0.0001$.

\paragraph{Names and sigla.} The target convention decides whether initials are reproduced as the scribe wrote them. Fine-tuned on sigla-as-written targets, the hybrid lowers the WER on the \hnChangedLines{} affected test lines, against the sigla reference, from \hnEWerCh\% to \hnVWerCh\% (CER \hnECerCh\% $\to$ \hnVCerCh\%; differences \hBsTgtChWerD{} [\hBsTgtChWerLo, \hBsTgtChWerHi] and \hBsTgtChCerD{} [\hBsTgtChCerLo, \hBsTgtChCerHi]); on all lines the WER goes from \hnEWerSig\% to \hnVWerSig\% and on names from \hnENames\% to \hnVNames\%. A named-entity head trained jointly (the \emph{NER head}) gives \hnNWerCh\% on the affected lines (CER \hnNCerCh\%), \hBsHeadChWerD{} [\hBsHeadChWerLo, \hBsHeadChWerHi] against the model without the head: with 31 lines this is not distinguishable from zero, and on all lines the two models are equal (\hnVWerSig\% and \hnNWerSig\% WER, \hnVWerExp\% and \hnNWerExp\% against the expanded transcription). $\alpha$ (\hnNAlpha\%), $\beta$ (\hnNBeta\%) and clause-final WER (\hnNFinal\% vs \hnEFinal\%) change little, and against the expanded transcription the sigla targets cost a little (WER \hnEWerExp\% $\to$ \hnVWerExp\%, CER \hnECerExp\% $\to$ \hnVCerExp\%), since the model no longer writes letters absent from the page. The constrained decoding never fired (\hnTagged{} name words, no siglum in any 5-best list). On silver v2 with R2L and sigla-guarded LM rescoring, the NER hybrid reaches \hWkNbestWer\% WER (expanded) and \hWkNbestWerSig\% (sigla reference; $\Delta$ \hBsNerWerD{} and \hBsNerSigWerD{}, $p<0.0001$; Table~\ref{tab:nersigla}), but on the sigla lines it does not differ significantly from the first NER hybrid (WER \hWkNbestWerCh\% vs \hnNWerCh\%, CI [\hBsNerChWerLo, \hBsNerChWerHi]). There the guard lowers the WER of LM rescoring from \hWkNuWerCh\% to \hWkNbWerCh\% (CER \hWkNuCerCh\% $\to$ \hWkNbCerCh\%); guarding only initials does the same at less cost overall (\hWkNsWer\% vs \hWkNbWer\% WER). Name WER stays near 30\%.

\section{Discussion and Conclusion}

Expert transcriptions without line breaks become training data at no annotation cost; data volume explains most of the accuracy, and re-alignment with the recognizer's own readings adds lines the graphemic reading could not place. An attention decoder read in both directions repairs clause-final inflections, and targets that keep initials as written make the output follow the scribe at a small cost against the edited text (a named-entity head adds no measurable gain). The space-free modular route we tested, a variant of \citet{camps2021}, loses once the recognizer has enough data to learn word spacing \citep{hasenohr1998}; their diplomatic chain remains untested here. The output convention is that of the targets and should be fixed before training: abbreviations expanded, name initials as written. A measuring recognizer must be deterministic; unmasked convolutions and a beam search that stops whole batches both make the output depend on the batch. The remaining errors sit at clause ends and in names, so machine-read registers need calibrated review and are a starting point, not a substitute for an edition \citep{schabel2023,burgers2022}.

\section*{Limitations}
All results come from one register volume and held-out pages of the same hands, and the test pages are all versos (the development pages all rectos; Section~2). Most models were trained with a single seed, so differences of a few tenths of a point may be within run-to-run variation. For the key model (hybrid, silver v2 + R2L) a second seed confirms the WER (\hWkRtlWer\% for both seeds, CER \hWkRtlCer\% and \hWkSeedCer\%, against \hHybWer\% WER for the first hybrid). The re-alignment of silver v2 used readings of the training pages decoded before the beam search was made batch-invariant; they only select spans of the expert transcription. Our permutation decoder is a simplified implementation, and masked-image pretraining used only the unlabeled lines of the training pages, because the unused pages of the volume are kept for calibration. The comparison with earlier work uses published figures on other test sets and metrics. The diplomatic variant of the modular route (diplomatic recognition followed by a neural expander) needs diplomatic ground truth, which does not yet exist for these registers. The NER head is trained and evaluated on rule-derived labels, since no named-entity annotation of the registers exists; its $F_1$ measures agreement with that rule, not with an expert. The gain of the head over sigla-as-written targets alone is small and not significant on the 31 affected lines. The headline figures cover the lines that align with the transcription (Section~3).

\section*{Author Contributions}
\textbf{Tamás Kovács}: Conceptualization, Methodology, Software, Validation, Formal analysis, Investigation, Data curation, Visualization, Writing -- original draft, Writing -- review \& editing. \\
\textbf{Daniel Luger}: Resources, Data curation, Writing -- review \& editing.

\bibliography{refs}

\appendix
\section{Training Details}\label{app:training}
\textbf{B} fine-tunes CATMuS Medieval~1.5 with \texttt{ketos} (Kraken~7.1.1): output layer resized to the union of character sets, early stopping on development accuracy with lag~3. The best model was reached after epoch \hBbest{} of \hBstages{} (development character accuracy \hBacc\%; about 11 minutes per epoch). \textbf{C}: line images cropped from the segmentation polygons, height 64 pixels; four convolutional blocks (convolution, batch normalization, SiLU, pooling; height $64\to4$, width $/4$); linear projection to $d=256$ with sinusoidal position encoding; eight Conformer blocks (macaron feed-forward, four heads, depthwise convolution kernel~15); CTC over the 64 characters of the targets; greedy decoding. Training: AdamW (learning rate $10^{-3}$, weight decay 0.01), two epochs of warm-up and cosine decay, width-bucketed batches of 24, bfloat16 mixed precision, gradient clipping at 5, width jitter $\pm10\%$ and contrast/brightness augmentation; early stopping after six epochs without improvement of the development CER, at most 60 epochs. C trained for \hCepochs{} epochs (best \hCbest, development CER \hCdev\%; \hCsec{}\,s per epoch on one RTX~A5000). The space-free model used the same setting (\hSFepochs{} epochs). \textbf{LM fusion}: prefix beam width 12; score $\log P_{\text{CTC}}+w\log P_{\text{LM}}+b$ per completed word, $w=0.5$ and $b$ chosen on development lines. \textbf{Segmenter}: character embedding (64) and upper-case flag, convolution (kernel~5), two-layer bidirectional LSTM (256), trained on 160-character windows; S3 weights the bigram model by 0.25, chosen on development lines. \textbf{Masked C}: as C, with the padded columns set to zero after every front-end block, padded frames set to zero at the entry of every Conformer block and before its depthwise convolution, and batch normalization computed over valid positions only (60 epochs; development CER 5.92\%). \textbf{Attention decoders}: four pre-norm Transformer decoder layers ($d=256$, four heads) over the masked encoder, character targets with end symbol, loss 0.3\,CTC $+$ 0.7 cross-entropy, early stopping on the development CER of greedy attention decoding (about 80\,s per epoch, 16.8\,M parameters). Hybrid: 5-best attention beam, hypotheses rescored by $0.3\log P_{\text{CTC}}+0.7\log P_{\text{att}}$. Copy gate: $p=g\,p_{\text{gen}}+(1-g)\sum_t a_t\,p_{\text{CTC}}(\cdot\mid t)$, with $a_t$ the cross-attention of the last layer and $g$ a sigmoid gate on the decoder state and the attended context. Permutation decoder: position queries attend to the context tokens allowed by six permutations per batch (left-to-right, right-to-left, four random); decoding left to right with one cloze refinement. Diplomatic head: CTC targets are the CATMuS readings of the silver lines (88 characters). \textbf{Silver v2 and its models} use the recipe of C and the hybrid, trained for all 60 epochs, with the development-best epoch kept. In the search over silver-v2 recipes (architecture and training variants of C and of the hybrid), a candidate run was stopped at epoch~12 if its development CER exceeded by more than 1.5 points the epoch-12 development CER of the original run of its kind (C: 7.18\%, hybrid: 8.41\%). Silver v2 was chosen among realignments with C, the hybrid or their union, at distance $\le$0.25 or $\le$0.20, with or without the hyphenation cut. \textbf{LM fusion of C}: prefix beam width 32, weights of the character and BPE models (0.7, 0.5) and a word bonus chosen on the development lines. \textbf{LM rescoring}: 16-best attention beam, score $=$ joint score $+\,a\log P_{\text{char}}/|h|+b\log P_{\text{BPE}}/\text{words}+c\log P_{\text{word}}/\text{words}$, with $(a,b,c)$ found by grid and coordinate search on the development lines. Minimum-Bayes-risk decoding on the same lists was not better on the development lines (WER \hWkMbrDevWer\% against \hWkBestDevWer\%). \textbf{NER on silver v2}: the NER fine-tuning recipe of Section~3 applied to the R2L hybrid; \emph{MWER} adds minimum word error rate training (8-best, learning rate $2\cdot10^{-5}$). Its n-best lists during training came from the original beam rule; all evaluation uses the fixed one. \textbf{Masked-image pretraining}: all 15,282 segmented lines of the training pages; 40\% of the width hidden in 16-pixel spans; a linear head reconstructs the $64\times4$ pixel block of every frame that overlaps a hidden column (mean squared error, 30 epochs), then CTC fine-tuning as for C. \textbf{Word error probability}: minimal CTC posterior of the word, isotonic calibration \citep{zadrozny2002transforming} on development words, separately inside and outside a lexicon; ECE \hEce\%.

\paragraph{Variants that did not help.} The table below lists the variants that did not improve the test WER over their reference (39 test pages, reference $\to$ variant, basis of Table~\ref{tab:variants}; hybrids decoded with the fixed beam). Vertical attention pooling of a 96-pixel front-end diverged twice and was stopped at epoch~12 (development CER above 70\%). Averaging the best checkpoints of a run (model soup) gained at most 0.1 points of development CER.

\noindent{\centering\small\setlength{\tabcolsep}{4pt}\resizebox{\columnwidth}{!}{\begin{tabular}{llrr}
\toprule
Variant & reference & WER & CER \\
 & & \% & \% \\
\midrule
auxiliary CTC on the CNN output & masked C & 14.3 $\to$ 14.6 & 6.9 $\to$ 6.8 \\
photometric/elastic augmentation & masked C & 14.3 $\to$ 14.3 & 6.9 $\to$ 7.0 \\
line height 96 & masked C & 14.3 $\to$ 14.9 & 6.9 $\to$ 6.9 \\
clause-final/name-weighted CE & hybrid & 12.5 $\to$ 12.9 & 6.4 $\to$ 6.7 \\
model soup (4 checkpoints) & hybrid v2 + R2L & 10.6 $\to$ 10.6 & 5.1 $\to$ 5.1 \\
\bottomrule
\end{tabular}
}\par}

\section{Corpus and Silver Lines}\label{app:data}
\begin{table}[h]
\centering\small\setlength{\tabcolsep}{3pt}
\resizebox{\columnwidth}{!}{\begin{tabular}{lrrrrr}
\toprule
 & entries & words & lines & accepted & silver \\
 & & & & \% & lines \\
\midrule
Year 1 (Luger), fol.~1--130 & 535 & 132{,}943 & 12{,}166 & 83.4 & 8{,}556 \\
Year 2, part (Singh Bottanova), fol.~137--203 & 325 & 70{,}876 & 6{,}966 & 80.8 & 4{,}566 \\
\midrule
train pages (315) & & & 15{,}282 & 82.5 & 10{,}525 \\
dev pages (40) & & & 1{,}846 & 84.8 & 1{,}313 \\
test pages (39) & & & 2{,}004 & 79.3 & 1{,}284 \\
\bottomrule
\end{tabular}}
\caption{Transcriptions, recognized lines, accepted lines (distance $\le0.35$) and silver lines ($\ge15$ characters, distance $\le0.25$), by transcription and by split.}
\label{tab:corpus}
\end{table}

\section{Error Bases}\label{app:bases}
Line CER pools all accepted lines of any length; compared-text CER (main text) uses accepted lines of at least 15 characters, averaged over pages; page CER compares all recognized lines of a page's main text block, accepted or not, with the reference span of the block, so it also counts marginalia, rubrics and reading-order errors (on fol.~28v, \hPageTwentyEight\% page CER against \hCmpTwentyEight\% compared-text CER). C's line CER (\hCline\%) is higher than its compared-text CER mainly because of short lines (\hCfifteen\% on lines $\ge$15 characters).
\begin{table}[h]
\centering\small\setlength{\tabcolsep}{3pt}
\resizebox{\columnwidth}{!}{\begin{tabular}{lrrrrr}
\toprule
Recognizer & line CER & compared CER & WER & page CER & aligned \\
 & \% & \% & \% & \% & \% \\
\midrule
A CATMuS Medieval 1.5 & 19.5 & 18.9 & 54.3 & 26.4 & 79 \\
A$'$ = A + rule expansion & 18.0 & 17.5 & 48.1 & -- & 80 \\
B CATMuS fine-tuned & 8.5 & 8.3 & 18.1 & 12.2 & 78 \\
C Conformer-CTC & 7.8 & 7.1 & 14.5 & 15.9 & 89 \\
C+G (C with Gaussian logit head) & 7.9 & 7.3 & 15.4 & 14.9 & 88 \\
Hybrid CTC/attention & 6.7 & 6.4 & 12.2 & -- & 89 \\
Hybrid + NER head, sigla as written & 7.1 & 6.8 & 12.6 & -- & 89 \\
Hybrid, re-aligned, R2L + BPE + LM & 4.9 & 4.7 & 8.5 & -- & 87 \\
Hybrid + NER head, re-aligned, R2L + LM & 5.3 & 5.1 & 9.5 & -- & 89 \\
\bottomrule
\end{tabular}}
\caption{Recognizers on three error bases (\%); WER with words aligned per page. A$'$: A with rule-based expansion of the CATMuS abbreviation signs. Aligned: share of the \hTestLines{} test lines accepted by the alignment (distance $\le0.35$); line, compared CER and WER use the accepted lines with at least 15 reference characters (compared CER: pages averaged). C+G: C with a heteroscedastic Gaussian logit head and a latent Gaussian bottleneck \citep{kendall2017}, trained with Monte Carlo CTC (8 samples, KL weight $10^{-3}$) to obtain per-frame uncertainty; it is no better than C and is not used elsewhere. The hybrids are scored against the expanded transcription (the re-aligned ones as in Table~\ref{tab:prior}); their page CER is not computed (--).}
\label{tab:bases}
\end{table}

\section{Learning Curve}\label{app:curve}
\begin{table}[h]
\centering\small
\begin{tabular}{rrrrrr}
\toprule
data & train lines & epochs & dev CER & CER & WER \\
 & & & \% & \% & \% \\
\midrule
10\% & 1{,}052 & 60 & 10.7 & 11.5 & 28.9 \\
25\% & 2{,}631 & 50 & 8.1 & 9.2 & 20.6 \\
50\% & 5{,}262 & 60 & 6.9 & 8.0 & 17.0 \\
100\% & 10{,}525 & 57 & 5.9 & 7.1 & 14.7 \\
\bottomrule
\end{tabular}
\caption{C trained on random samples of the silver training lines: epochs to early stopping, best development CER, and test CER and WER (words aligned per line).}
\label{tab:curve}
\end{table}
\begin{figure*}[h]
\centering
\includegraphics[width=\textwidth]{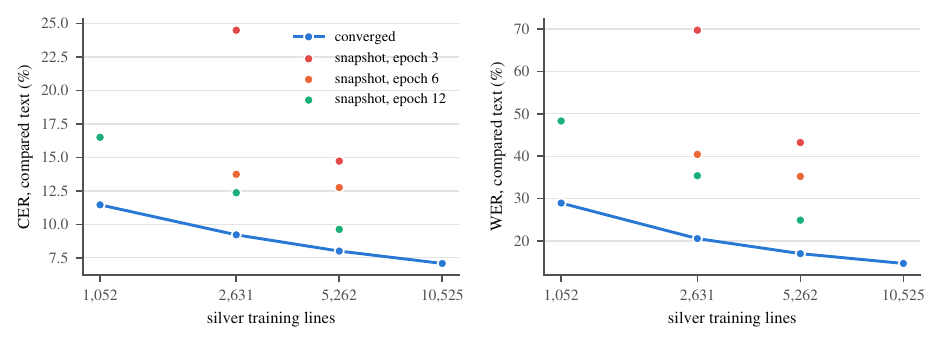}
\caption{CER and WER of C against the number of silver training lines: converged runs and early snapshots after epochs 3, 6 and 12.}
\label{fig:curve}
\end{figure*}

\section{Errors by Position}\label{app:errors}
Clause ends are tokens of the reference ending in \emph{. ? ; !}, excluding abbreviation points (initials, numerals, \emph{Kal.}, \emph{Id.} and similar) and register truncations (\emph{etc.}, \emph{usque}). A word counts as abbreviated if the CATMuS reading of the same position contains an abbreviation sign.
\begin{table}[h]
\centering\small\setlength{\tabcolsep}{3pt}
\resizebox{\columnwidth}{!}{\begin{tabular}{lrrrrrr}
\toprule
 & \multicolumn{4}{c}{C} & hybrid & + NER \\
\cmidrule(lr){2-5}
Position (reference clause ends) & words & WER & abbreviated & in full & WER & WER \\
 & & \% & \% & \% & \% & \% \\
\midrule
clause-final word & 619 & 22.3 & 25.4 & 19.6 & 16.4 & 17.3 \\
penultimate word & 600 & 19.5 & 21.1 & 18.1 & 16.8 & 16.2 \\
3rd--4th word from the end & 1{,}160 & 20.3 & 16.0 & 25.0 & 17.1 & 17.2 \\
all other words & 17{,}974 & 13.9 & 13.6 & 14.1 & 11.9 & 12.3 \\
\midrule
all words & 20{,}353 & 14.7 & 14.4 & 15.0 & 12.5 & 12.8 \\
\bottomrule
\end{tabular}}
\caption{Word error by position relative to clause ends: C, for words abbreviated on the page and words written in full, and the hybrid without and with the NER head (sigla as written), against the expanded transcription (words aligned per line together with the diplomatic reading, which gives \hErrFinal\% for clause-final words against \hCFinal\% in Table~\ref{tab:variants}).}
\label{tab:errors}
\end{table}

\section{Names and Sigla}\label{app:names}
\begin{table*}[t]
\centering\small\setlength{\tabcolsep}{4pt}
\begin{tabular}{lrrrrrrrrr}
\toprule
 & \multicolumn{2}{c}{expanded ref.} & \multicolumn{2}{c}{sigla ref., all} & \multicolumn{2}{c}{sigla ref., 31 lines} & names & & \\
\cmidrule(lr){2-3}\cmidrule(lr){4-5}\cmidrule(lr){6-7}
Recognizer & WER & CER & WER & CER & WER & CER & WER & $\alpha$ & $\beta$ \\
\midrule
hybrid, names expanded (E10) & 12.5 & 6.4 & 8.9 & 4.5 & 13.3 & 10.1 & 32.1 & 8.1 & 10.1 \\
hybrid, sigla as written & 12.8 & 6.8 & 8.7 & 4.3 & 9.2 & 5.0 & 29.6 & 7.8 & 10.5 \\
hybrid, sigla + NER head & 12.8 & 6.8 & 8.7 & 4.3 & 8.2 & 4.9 & 29.4 & 7.2 & 10.6 \\
\midrule
hybrid, silver v2 + R2L + NER head & 10.5 & 5.2 & 8.1 & 4.0 & 9.2 & 5.9 & 31.0 & 11.7 & 5.0 \\
\quad + LM rescoring, sigla guard & 9.8 & 5.1 & 7.2 & 3.8 & 7.7 & 5.2 & 31.7 & 11.1 & 4.9 \\
\quad + MWER + LM rescoring, sigla guard & 9.6 & 5.0 & 7.1 & 3.7 & 7.5 & 5.6 & 30.8 & 9.4 & 4.0 \\
\bottomrule
\end{tabular}

\caption{Hybrid recognizers and name initials on the 39 test pages. Expanded ref.: the common basis of Table~\ref{tab:variants} (names expanded by the editor); sigla ref.: the transcription with name initials as written, on all test lines and on the lines the rule changes. Names: WER on words the name rule selects in the sigla reference. $\alpha$, $\beta$ on the common basis. Below the line: the NER fine-tuning applied to the R2L hybrid of silver v2, alone, with LM rescoring under the sigla guard, and after MWER training with the same rescoring. Without the guard, LM rescoring gives \hWkNuWerCh\% / \hWkNuCerCh\% WER/CER on the \hnChangedLines{} lines (MWER: \hWkNmWerCh\% / \hWkNmCerCh\%).}
\label{tab:nersigla}
\end{table*}

\textbf{Name rule and training.} A name is a capitalized word that does not open a clause, is not a Collatinus form \citep{ouvrard_collatinus} and is not a chancery, calendar or ecclesiastical title (\emph{Kal.}, \emph{Ianuarii}, \emph{episcopo}, \emph{Domino}, Roman numerals); place adjectives in \emph{-ensis} count as places, other names as persons, and a word is tagged when most of its letters are. The text targets carry no tags, so the output sequence and its length are unchanged. Both models start from the hybrid and train for six epochs at a learning rate of $10^{-4}$ without early stopping: early stopping on the development CER, which the sigla rule changes on only about 2\% of the lines, kept the starting model. The head agrees with the rule labels at $F_1=\hnHeadFone$ (P \hnHeadP, R \hnHeadR) when it reads the reference text of the development lines, and at $F_1=\hnEtoEF$ on its own recognized text (relaxed match, P \hnEtoEP, R \hnEtoER).

The head is evaluated against labels from the same rule that trained it, so its $F_1$ measures how well it reproduces the rule from the decoder states, not agreement with an expert. As entity suggestions for review, the head and the rule can be combined. On the development lines (relaxed match, recognized text) the rule alone reaches $F_1=\hnFilterF$, the head \hnEtoEF, their union \hnOrF{}, and their intersection the highest precision (\hnAndP). The rule's advantage is partly circular, since the reference labels come from the rule applied to the expert text.

\section{Reproducibility}\label{app:repro}
The original C depends on the inference batch: its key-padding mask covers self-attention, but the convolutions and batch normalization also see the padded frames. Decoding the test lines in different batch layouts changes up to \hPadMax{} of 2,004 lines and moves WER and $\alpha$ (Table~\ref{tab:padding}). The CTC decoding already used the subsampled sequence lengths, and inference already grouped lines of similar width. Masking the padded positions in every convolutional and normalization layer removes the dependence, even with the weights of C and without retraining (at most \hPadMaskInf{} lines change, floating-point noise). A model trained with mask propagation gives identical metrics for every batch layout. All figures for C and its checkpoints in this paper use masked, batch-invariant decoding of the trained weights. Unmasked decoding left the WER unchanged, because the padding only altered words that were wrong already, but it moved CER and $\alpha$.

The attention beam search of the hybrids had a second batch dependence. A batch stopped only when every line had $K$ finished hypotheses, so a line that finished early kept collecting candidates while longer lines of its batch were decoded, and its $K$-best list and joint choice depended on its neighbors (the permutation decoder's refinement step likewise saw the tokens generated after the end symbol). With this rule, \hPadHybOld{} of the 2,004 test lines of the first hybrid change between batch layouts, and the WER moves between \hPadHybOldWerLo\% and \hPadHybOldWerHi\% (Table~\ref{tab:padding}). Stopping each line at its own $K$ hypotheses, as when it is decoded alone, leaves at most \hPadHybFixed{} lines changed (\hPadBpeFixed{} for the best single hybrid), floating-point differences between kernels for different batch shapes. Every hybrid figure in this paper, including the development lists used for selection, comes from the fixed beam search. Code, metrics and the script that generates every number of this paper are available from the authors; the page images are subject to the rights of the Vatican Apostolic Archive.
\begin{table}[h]
\centering\small\setlength{\tabcolsep}{3pt}
\resizebox{\columnwidth}{!}{\begin{tabular}{lrrrrrr}
\toprule
Model & batch & lines changed & CER & WER & $\alpha$ & $\beta$ \\
 & & vs.\ batch 1 & \% & \% & \% & \% \\
\midrule
C as trained & 1 & 0 & 7.08 & 14.48 & 10.5 & 13.6 \\
 & 16 & 465 & 7.01 & 14.39 & 11.2 & 13.2 \\
 & 32 & 598 & 7.14 & 14.48 & 12.1 & 13.4 \\
 & 32, random & 810 & 7.08 & 14.44 & 9.5 & 13.2 \\
\midrule
C weights, masked inference & 1 & 0 & 7.09 & 14.49 & 10.5 & 13.6 \\
 & 16 & 5 & 7.10 & 14.50 & 10.5 & 13.8 \\
 & 32 & 7 & 7.08 & 14.48 & 10.5 & 13.6 \\
 & 32, random & 8 & 7.10 & 14.50 & 10.5 & 13.8 \\
\midrule
C trained with mask propagation & 1 & 0 & 6.91 & 14.05 & 8.4 & 11.0 \\
 & 16 & 9 & 6.92 & 14.05 & 8.4 & 11.0 \\
 & 32 & 8 & 6.91 & 14.05 & 8.4 & 11.0 \\
 & 32, random & 9 & 6.91 & 14.05 & 8.4 & 11.0 \\
\midrule
Hybrid, original beam stop & 1 & 0 & 6.42 & 12.25 & 8.1 & 10.1 \\
 & 16 & 270 & 6.27 & 12.11 & 8.9 & 9.6 \\
 & 32 & 298 & 6.27 & 12.09 & 9.0 & 9.4 \\
 & 32, random & 402 & 5.95 & 11.62 & 9.0 & 9.5 \\
\midrule
Hybrid, fixed beam stop & 1 & 0 & 6.42 & 12.25 & 8.1 & 10.1 \\
 & 16 & 3 & 6.41 & 12.24 & 8.1 & 10.1 \\
 & 32 & 3 & 6.41 & 12.24 & 8.1 & 10.1 \\
 & 32, random & 3 & 6.41 & 12.24 & 8.1 & 10.1 \\
\midrule
Hybrid v2 + R2L + BPE, fixed & 1 & 0 & 4.75 & 9.23 & 8.3 & 6.8 \\
 & 16 & 0 & 4.75 & 9.23 & 8.3 & 6.8 \\
 & 32 & 1 & 4.76 & 9.23 & 8.3 & 6.8 \\
 & 32, random & 0 & 4.75 & 9.23 & 8.3 & 6.8 \\
\bottomrule
\end{tabular}}
\caption{Batch dependence of the output: lines changed against decoding one line at a time, and the metrics of each batch layout (words aligned per page). Batches are grouped by line width unless marked random. Hybrids: joint 5-best decoding under the original stopping rule of the beam search (a batch stops when all its lines are finished) and the fixed rule (each line stops on its own).}
\label{tab:padding}
\end{table}

\end{document}